\documentclass[11pt]{article}
\usepackage[a4paper, total={6in, 8in}]{geometry}
\usepackage[english]{babel}
\usepackage{amsmath, amsfonts, amssymb, amsthm}
\usepackage{graphicx}
\usepackage[colorlinks=true, allcolors=blue]{hyperref}
\usepackage{algorithmic}
\usepackage{algorithm}
\usepackage{array}
\usepackage[caption=false,font=normalsize,labelfont=sf,textfont=sf]{subfig}
\usepackage{textcomp}
\usepackage{stfloats}
\usepackage{url}
\usepackage{verbatim}
\usepackage{booktabs}
\usepackage{orcidlink}
\usepackage{caption3} 
\DeclareCaptionOption{parskip}[]{} 
\usepackage[footnotesize]{caption}

\providecommand{\keywords}[1]
{
  \small	
  \textit{{Keywords --}} #1
}

\title{Reconstructing Movement from Sparse Samples: Enhanced Spatio-Temporal Matching Strategies for Low-Frequency Data}

\author{Ali Yousefian$^{1}$, Arianna Burzacchi$^{12*}$\orcidlink{0000-0001-8284-4909}, Simone Vantini$^{1}$\orcidlink{0000-0001-8255-5306}  \\
    \small $^{1}$ MOX, Department of Mathematics, Politecnico di Milano, \\ \small Piazza Leonardo da Vinci 32, 20133, Milan, Italy\\
    \small $^{2}$ Modeling and Simulation of Socio-Technical Systems, \\
    \small Fondazione Bruno Kessler, Via Sommarive 18, 38123, Trento, Italy \\
    \small $^{*}$ Corresponding author: aburzacchi@fbk.eu \\
}

\date{}

\begin{document}
\theoremstyle{definition}
\newtheorem*{definition}{Definition}
\theoremstyle{remark}
\newtheorem*{remark}{Remark}

\maketitle

\begin{abstract}
This paper explores potential improvements to the Spatial-Temporal Matching algorithm for aligning the GPS trajectories to road networks. While this algorithm is effective, it presents some limitations in the computational efficiency and the accuracy of the results, especially in dense environments with relatively high sampling intervals. To address this, the paper proposes four modifications to the original algorithm: a dynamic buffer, an adaptive observation probability, a redesigned temporal scoring function, and a behavioral analysis to account for the historical mobility patterns. 
The enhancements are assessed using real-world data from the urban area of Milan, and through newly defined evaluation metrics to be applied in the absence of ground truth. The results of the experiment show significant improvements in performance efficiency and path quality across various metrics. 
\end{abstract}

\keywords{Data analytics and data science,
Traffic networks,
Map Matching, 
GPS data}

\section{Introduction}\label{sec:introduction}

The widespread use and availability of Global Positioning System (GPS) technologies now enables the continuous monitoring of vehicles and people across both time and space. 
However, due to intrinsic errors and inaccuracies in positioning systems, raw trajectory data often contain noise, missing segments, and spatial misalignment. 
Several factors may negatively affect the accuracy of the individual GPS points recorded, such as signal blockage in dense urban environments and inconsistency of GPS receivers. As a consequence, the recorded trajectory may inevitably deviate from the true path taken on the road network \cite{huang2021survey}.
In addition, by nature, GPS data provide only a partial view of real-world movement: continuous motion is observed only at discrete time intervals, capturing therefore just a subset of the true trajectory. This limitation becomes more pronounced when the data are sampled at low frequency, resulting in large temporal gaps between consecutive observations. In such cases, important details of the trajectory may be lost, making it difficult to precisely reconstruct the true travel path \cite{rakhman2023gps}.

To address these limitations, a preprocessing stage is required to facilitate better interpretation of the raw location data, with one of the most essential steps being Map Matching \cite{survey2020}. 
The technique is utilized to match the potentially noisy GPS observations of spatial trajectories with their real position on the underlying transportation network, identifying the correct road segment on which a moving vehicle or user travels, and their position on that segment \cite{quddus2007current}. It acts as a fundamental step in various location-based applications such as navigation, tracking of vehicles, traffic monitoring, and intelligent transport systems \cite{survey2020, yuan2010interactive}. 

The diversity of road geometries and GPS characteristics has led to a vast landscape of Map Matching algorithms. Literature traditionally categorizes these approaches based on the information they exploit, such as geometric, topological, and probabilistic features \cite{huang2021survey, quddus2007current, yuan2010interactive}, or the underlying computational model, ranging from similarity-based to state-transition and learning-based frameworks \cite{survey2020, Yuan2021Survey}.
However, the effectiveness of these methods is strictly dependent on the sampling frequency of the trajectory data. While geometric and topological methods perform well with high-density data, they often fail in low-frequency scenarios, where the sparse connectivity between points introduces significant ambiguity. 
Low-frequency data has longer time intervals between two consecutive GPS observations. As cited in \cite{huang2021survey}, intervals greater than 30 seconds are considered low-frequency data, and above 2 minutes are ultra-low-frequency data. Statistical analysis in \cite{yuan2010interactive} has shown that in practice there is a large amount of low sampling rate GPS data, typically above the two minute threshold. Low-frequency data make it harder to predict the real route that a moving object has taken since the distance between neighboring GPS points is higher and there might be several different possible paths between them. 


In this context, one widely cited method for low-frequency GPS data is the ST-Matching, proposed by Lou et al. in 2009 \cite{lou}. This algorithm is a weighted graph technique, i.e., a specific case of state transition models \cite{survey2020}. Indeed, it forms the matched path based on the weights of a candidate graph, whose nodes represents the point locations on the road map that are candidates for being the true position of the GPS record, and whose edges connect pairs of consecutive points and are weighted based on spatial proximity and a temporal function. As a result, the target becomes finding the longest path on this weighted graph, i.e., the sequence of candidates that maximizes the total weight utility. Despite its effectiveness, the method struggles with complex road topologies, noisy GPS data, and scalability issues when processing large datasets. 

The goal of our work is to investigate how the ST-Matching algorithm can be improved both in terms of its performance efficiency and the quality of its results. To address this, a set of modified versions of the method is proposed, each designed to overcome specific limitations of the original implementation and enhance the overall performance. 
In particular, the modified ST-Matching is introduced, refining the original method by improving the candidate point selection method, the spatial analysis component, and the temporal component to better reflect realistic movements, travel time, and speed. Moving on to the realm of low-frequency data, we investigate the potential of adding a behavioral score based on historical data to the algorithm, leading to the design of a new behavioral score and the development of the STB-Matching method.

In addition, this paper presents an evaluation framework designed to assess the performance of the modified algorithms. Several metrics are proposed to examine the methods from different perspectives, including computational efficiency, matching accuracy, adherence to topological constraints, and execution speed. 
Throughout them, a comprehensive comparative analysis is performed to assess the improvements of the new methodologies with respect to the baseline method. The performance and results of thse algorithms are also compared to the baseline on low-frequency data. 
The proposed approach is validated through a case study in Milan, a key metropolitan area in Northern Italy. The results show clear improvements over the baseline ST-Matching implementation and highlight which modifications are most effective under various operating conditions.

The remainder of this paper is organized as follows. Section \ref{sec:methodology} describes the methodology in detail, including both the baseline and the proposed new algorithms. Section \ref{sec:case study} presents the case study application, including the urban area of interest, road network description, data source and structure, and the preprocessing steps used to prepare the dataset. It also presents and discusses the results of the experiment and compares the performance of the algorithms. In the end, Section \ref{sec:conclusion} derives a conclusion and points out potential directions for future work. 
The implementation of the algorithms and the analysis of the results in this paper are carried out using the programming language \texttt{Python} \cite{python}. All the code is open source and available in the GitHub repository \texttt{map-matching} \cite{github-repo-ali}.

\section{Methodology}
\label{sec:methodology}

\subsection{Prior Definitions}

\begin{definition}[GPS point]
A GPS point $p$ is defined as a combination of the variables of spatial coordinates and the timestamp. Formally, $p=(x,y,t)$, where $x$ and $y$ are the spatial coordinates, and $t$ is the timestamp. 
\end{definition}

\begin{definition}[GPS Trajectory]
A GPS trajectory $T$ consists of a sequence of points $p_1\rightarrow p_2\rightarrow \dotsi p_i\dotsi \rightarrow p_n$, ordered with respect to their timestamp ($t_{i+1} > t_{i}$ for $i \in \{1, 2, \dots, n-1\}$) and related to the movements of the same subject (user, vehicle).
\end{definition}

\begin{definition}[Road Network]
The network of the streets in an area which is defined by a directed graph $G(V, E)$ where $E$ is a set of edges representing street segments and $V$ is a set of vertices representing the intersections or terminal points of a segment. Each edge $e\in E$ usually has a range of features, such as \textit{id}, \textit{length}, \textit{travel speed}, and \textit{number of lanes}.
\end{definition}

\begin{definition}[Candidate Points]
For each point $p_i$ of a trajectory, a set of candidate points \{$c_i^1$, $c_i^2$, $\dots$, $c_i^{m_i}$\} consists of the closest points to $p_i$ projected on the road network edges within a certain search radius.
\end{definition}

\begin{definition}[Path]
Considering the the road network graph $G(V,E)$, the path $P_{ij}$ between two vertices $V_i$ and $V_j$ is a set of connected road segments in $E$, $e_1\rightarrow e_2 \rightarrow \dotsi \rightarrow e_k$, where $V_i$ is the starting point of the first edge $e_1$, and $V_j$ is the ending point of the last edge $e_k$.  
\end{definition}

\begin{definition}[Map Matching]
The problem of Map Matching can be described as finding the best candidate point $\tilde{c}_i$ for each $p_i$ of the trajectory $T$.
\end{definition}

\begin{definition}[Path Reconstruction]
The path reconstruction problem is defined as finding the best path $\tilde{P}=(\tilde{e}_1 \rightarrow ... \rightarrow \tilde{e}_k)$ on graph $G$ that matches the trajectory $T$ with its real path. Optimal candidates are located on these edges: $\forall i\in\{1,\dots,n \}\: \exists j\in\{1,\dots,k\} :\tilde{c}_i \in \tilde{e}_j$.
\end{definition}

\subsection{Overview of the Baseline Method}
\label{sec:overview}
While a comprehensive description of the ST-Matching algorithm can be found in the reference work \cite{lou}, in this section, a brief summary is presented. The ST-Matching algorithm consists of three main stages: (i) \textit{candidate preparation}; (ii) \textit{spatial and temporal analysis}; (iii) \textit{result matching}.


\begin{remark}
In the following formulas, let $i$ be the GPS point index, with $2\leq i \leq n$ per each trajectory $T$. Let $t$ and $s$ be the candidate indices of GPS points $i-1$ and $i$ respectively, with $1\leq t\leq m_{i-1}$ and $1\leq s\leq m_{i}$.
\end{remark}

\begin{figure}[tb]
    \centering
    \includegraphics[width=0.5\linewidth]{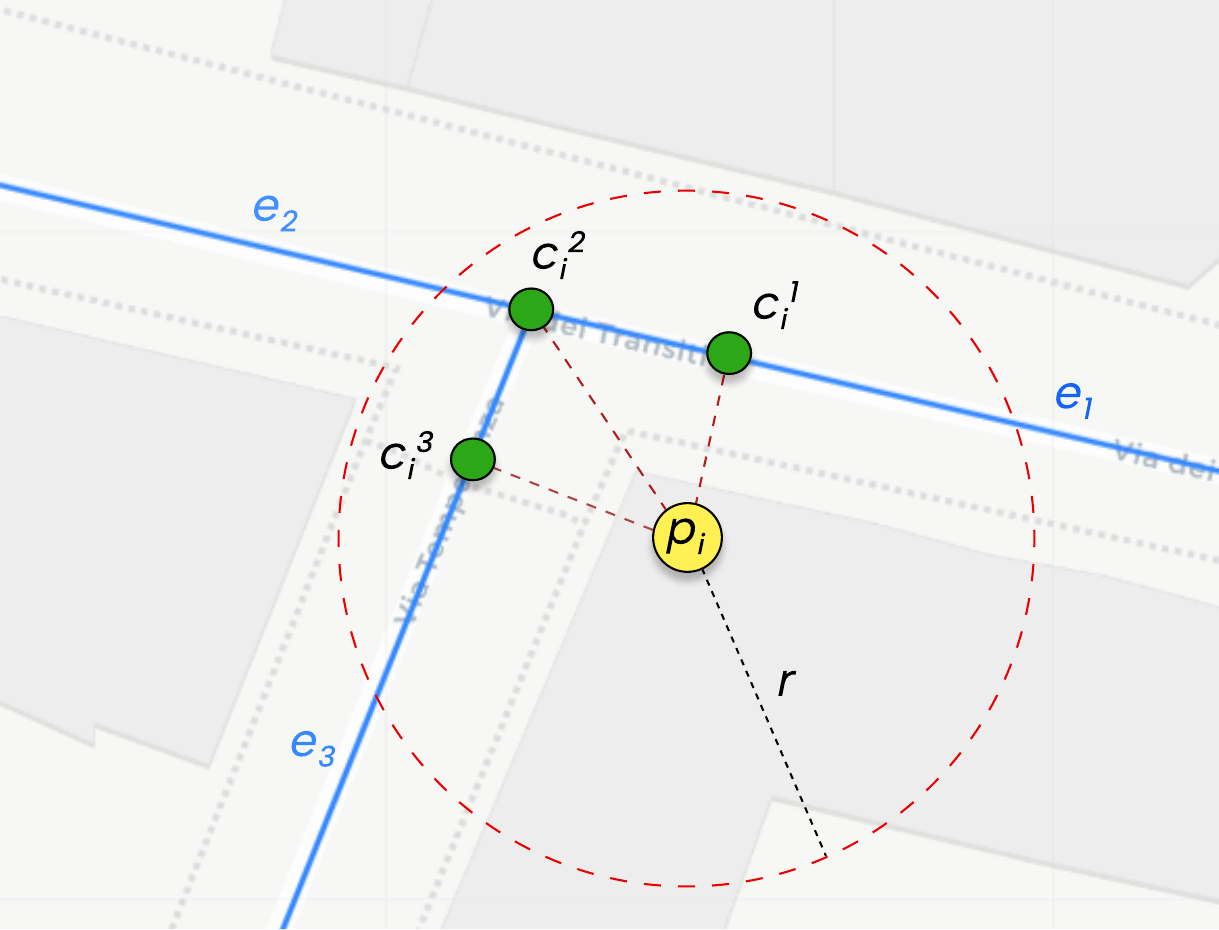}
    \caption{Retrieving candidate points and candidate edges for the GPS point $p_i$.}
    \label{fig:candidate selection}
\end{figure}

In the first phase, each GPS record is associated with a set of possible candidate points matched on the road network. Specifically, for each point $p_i$ in a trajectory $T$, first, a set of candidate segments is retrieved on the road network within a radius $r$. This range is fixed and is suggested to be $r=100$ meters \cite{lou}.
For each edge, if the projection of the point $p_i$ falls within its endpoints, the projection is selected as a candidate point. Otherwise, the endpoint which is closer to $p_i$ will be chosen as the candidate point of that segment. As a result for each $p_i$ there is a set of $m_i$ candidate points \{$c_i^1, c_i^2,\dots,c_i^{m_i}$\}. This procedure is illustrated in Figure \ref{fig:candidate selection}. After retrieving all candidate points $c_i^j$ for all sample points $p_i$, the problem of Map Matching turns into finding the best sequence of candidate points $\tilde{c}_1 \rightarrow \tilde{c}_2 \rightarrow\dotsi \tilde{c}_n$ that matches the trajectory $T$ with its real path. 

In the subsequent step, spatial and temporal analyses are performed to properly evaluate the candidate selection. 
The spatial analysis takes into account both geometric and topological aspects of the road network, where the former is expressed through the \textit{observation probability} and the latter is defined by the \textit{transmission score}. 

The observation probability measures the likelihood that GPS sample point $p_i$ corresponds to a candidate point $c_i^s$, given the Euclidean distance between them.
This likelihood assumes that the GPS error has a normal distribution, typically with a zero mean and a standard deviation ($\sigma$), which is set to $\sigma=20$ based on the empirical evaluations \cite{lou}. The observation probability is calculated using the probability density function of such a distribution as follows:
\begin{equation}
\label{eq:observation probability}
N(c_i^s) = \frac{1}{\sqrt{2\pi\sigma}} \, \exp\left(-\dfrac{\text{dist}_e(p_i,c_i^s)^2}{2\sigma^2}\right) ,
\end{equation}
where $\text{dist}(p_i, c_i^s)$ is the Euclidean distance between $p_i$ and $c_i^s$.

The observation probability handles each GPS point in isolation, not considering its neighboring points in the trajectory, and might select a spatially closer but contextually incorrect match. 
The transmission score helps account for the topological relationship between consecutive candidate points. It is defined as the likelihood that the true path between two candidate points follows the shortest path between them. It is computed as the ratio of the Euclidean distance between two consecutive GPS sample points and the shortest network distance between their corresponding candidate points:
\begin{equation}
\label{eq:transmission probability}
V(c_{i-1}^t \rightarrow c_{i}^s) = \frac{\text{dist}_e(p_{i-1}, p_i)}{\text{dist}_{n}(c_{i-1}^t ,\ c_{i}^s)},
\end{equation}
where \( \text{dist}_e(p_i, p_{i-1}) \) is the Euclidean distance between GPS points \( p_i \) and \( p_{i-1} \); \( \text{dist}_{n}(c_{i-1}^t ,\ c_{i}^s) \) is the length of the shortest path on the road network between candidate points \( c_{i-1}^t \) and \( c_i^s \).
The shortest path can be calculated using a variety of methods, such as the A* algorithm \cite{hart1968formal}, a widely used approach that employs a heuristic function to guide the search toward the goal. 

To combine geometric proximity and contextual consistency, the algorithm multiplies the observation probability by the transmission score to create the \textit{spatial score:}
\begin{equation}
\label{eq:spatial score}
F_\text{spatial}(c_{i-1}^t \rightarrow c_{i}^s) = N(c_{i}^s) \cdot V(c_{i-1}^t \rightarrow c_{i}^s).
\end{equation}
For any two consecutive sample points $p_{i-1}$ and $p_i$, a set of candidate points $\{c_{i-1}^t\}$ and $\{c_i^s\}$ is found, then each path between candidate pairs is scored using Equation \ref{eq:spatial score}.

The defined spatial analysis alone is not enough to estimate the true path among the candidate points: situations may arise in which different candidate paths are spatially plausible, even though their actual feasibility can differ significantly when realistic vehicle speeds are considered. The temporal analysis addresses this ambiguity by comparing the average speed between two consecutive GPS observations with the speed profiles of their candidate segments.
It does so by extracting the segment speed limits for each candidate path, matching them with a vector based on the observed average speed, and comparing the two via cosine similarity:
\begin{equation} 
\label{eq:temporal score}
F_\text{temporal}(c_{i-1}^t \rightarrow c_i^s) 
=\frac{
\sum_{u=1}^{k} (v_{lim,e_u} \; \cdot \; \bar{v}_{(i-1,t)\rightarrow(i,s)})
}{
\sqrt{\sum_{u=1}^{k} (v_{lim,e_u})^2} \cdot \sqrt{\sum_{u=1}^{k} \bar{v}_{(i-1,t)\rightarrow(i,s)}^2}
},
\end{equation}
where $v_{lim,e_u}$ is the speed limit in the road segment $e_u$, $\bar{v}_{(i-1, t)\rightarrow(i,s)}$ is the average speed of the shortest path between $c_{i-1}^t$ and $c_i^s$, and $[e_1 \,,\, \dots \,,\, e_k] \subseteq E$ is the list of road segments of this path.

The final ST-Score between consecutive candidate points, $c_{i-1}^t$ and $c_i^s$, is computed by multiplying the two functions of Equations \ref{eq:spatial score} and \ref{eq:temporal score}, obtaining:
\begin{equation}
\label{eq: final score}
    F_\text{ST}(c_{i-1}^t \rightarrow c_i^s)  = F_\text{spatial}(c_{i-1}^t \rightarrow c_i^s)  \cdot F_\text{temporal}(c_{i-1}^t \rightarrow c_i^s) .
\end{equation}

\begin{figure}[ht]
    \centering
    \includegraphics[width=0.4\linewidth]{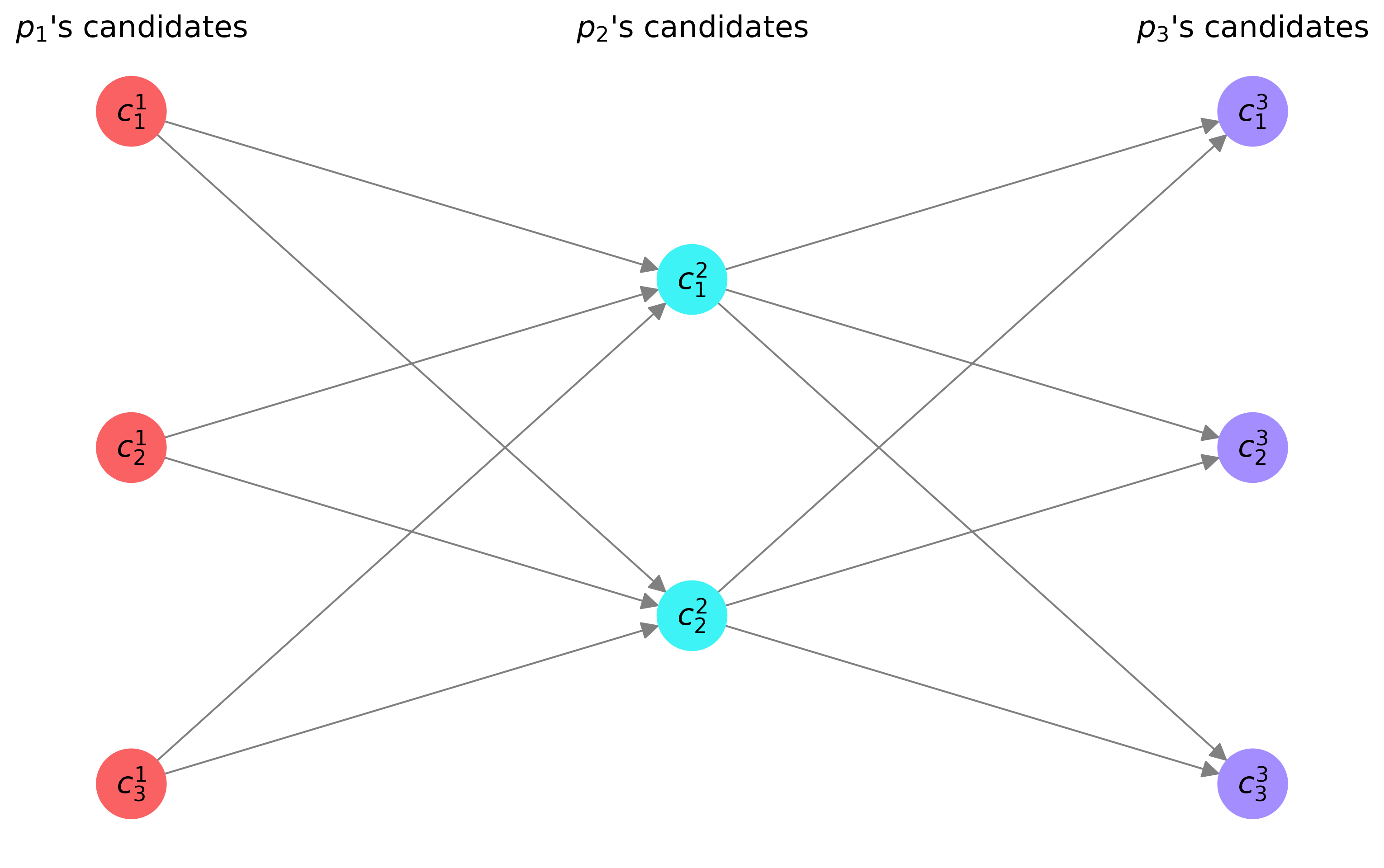}
    \caption{An example of a candidate graph for a 3-point trajectory.}
    \label{fig:candidate graph}
\end{figure}

After spatial and temporal analysis, for each trajectory $T$, it is possible to create a candidate graph. It is a Trellis graph \cite{ryan1993viterbi}, of which the set of nodes includes the candidate points for each GPS sample $p_i$, and the set of edges is made of links connecting candidates of consecutive points. Formally, the candidate graph is defined as $G_T =(V_T, E_T)$ with:
\begin{equation}
\begin{split}
    V_T =& \left\{ c_i^s, \quad \forall s=1, \dots, m_i \:\:\: \forall i=1, \dots, n  \right\}, \\
    E_T = &\left\{ (c_{i-1}^t \rightarrow c_i^s), \quad \forall t=1, \dots, m_{i-1} \:\: \forall s=1, \dots, m_i \:\: \forall i=1, \dots, n \right\}.
    \\
\end{split}    
\end{equation}
Moreover, each edge of the graph is characterized by the ST-Score given by the formulation in Equation \ref{eq: final score}. Figure \ref{fig:candidate graph} outlines an example of a candidate graph.

Now the problem of path reconstruction becomes finding the path in $G_T$ that maximizes the overall $F_\text{ST}$ score, or in other words, finding the longest path on this graph weighted by $F_\text{ST}$. This corresponds to identifying the sequence of candidates that, as a whole, offers the most plausible reconstruction of the underlying GPS trajectory.

\subsection{Proposed Modifications to ST-Matching}
While ST-Matching can be a robust baseline algorithm for Map Matching, there is still room for improvement in the way it achieves its goal and its performance efficiency. In this work four key modifications are suggested and investigated: \begin{enumerate}
    \item Dynamic candidate preparation, which affects the search area for the candidate points using a dynamic buffer and through GPS accuracy. It is discussed in Section \ref{sec:dynamic buffer};
    \item Dynamic observation probability, which modifies the first component of the spatial analysis again based on data accuracy, and is presented in Section \ref{sec:dynamic buffer};
    \item Redesigned temporal score function, which replaces entirely the temporal function of the original algorithm and is detailed in Section \ref{sec:temporal function};
    \item An added behavioral score component that is a completely new function introduced to that algorithm to capture the behavioral impact of the road network users according to the historical data. Section \ref{sec:behavioral} presents this component in details.
\end{enumerate}

\subsubsection{Dynamic Candidate Preparation and  Observation Probability}
\label{sec:dynamic buffer}
The original ST-Matching algorithm relies on several constraints that limit its flexibility. Specifically, the buffer radius $r$ for retrieving candidate points is fixed ($r=100$ meters), as suggested in \cite{lou}. While this approach, combined with a cap on the number of candidates ($m_i=5$), reduces the density of the candidate graph and hence overall runtime, it introduces a significant trade-off: it may exclude the ground-truth segment in complex environments or sparse sampling scenarios. Since the computational cost is directly proportional to the number of shortest-path searches between consecutive candidates, these rigid parameters are often necessary to maintain efficiency. Furthermore, the observation probability (Eq. \ref{eq:observation probability}) depends on a fixed standard deviation ($\sigma=20$ meters), which fails to account for varying GPS noise levels. These limitations highlight the need for a more adaptive approach to parameter selection.


In this paper, we leverage the GPS uncertainty feature, which is widely available to measure the quality of the records. GPS uncertainty, though it can be characterized in several different ways, in this approach is defined as the radius (in meters) of a circle around the recorded GPS point with a 68\% probability that the real point falls inside of it. This means the higher the value, the higher the uncertainty, the lower the accuracy and vice versa.

To avoid using fixed values or parameter tuning, we create \textit{dynamic buffer} and \textit{dynamic observation probability} based on the uncertainty feature of the dataset.  
%
First, the GPS uncertainty value is used to determine the buffer area within which the search for candidate points and road segments is performed. In this way, the search buffer scales dynamically with the uncertainty of the GPS measurement: a higher uncertainty expands the radius to capture all potential candidates, while a lower uncertainty narrows the search area, effectively reducing the number of candidate points and the overall computational load. In this work, the buffer value starts from the value of the GPS uncertainty, and if no candidate points are found within this radius, it is incremented by a step of 2 meters until a candidate point is found. A maximum radius, $r_{max}$, of 50 meters is selected based on preliminary tests. Unlike the 100-meter buffer suggested in \cite{lou}, which significantly increases processing time, our 50-meter threshold ensures computational feasibility. By dynamically excluding less plausible candidates, this approach aims to enhance both algorithmic efficiency and matching accuracy.

\begin{figure}[tb]
    \centering
    \includegraphics[width=0.5\linewidth]{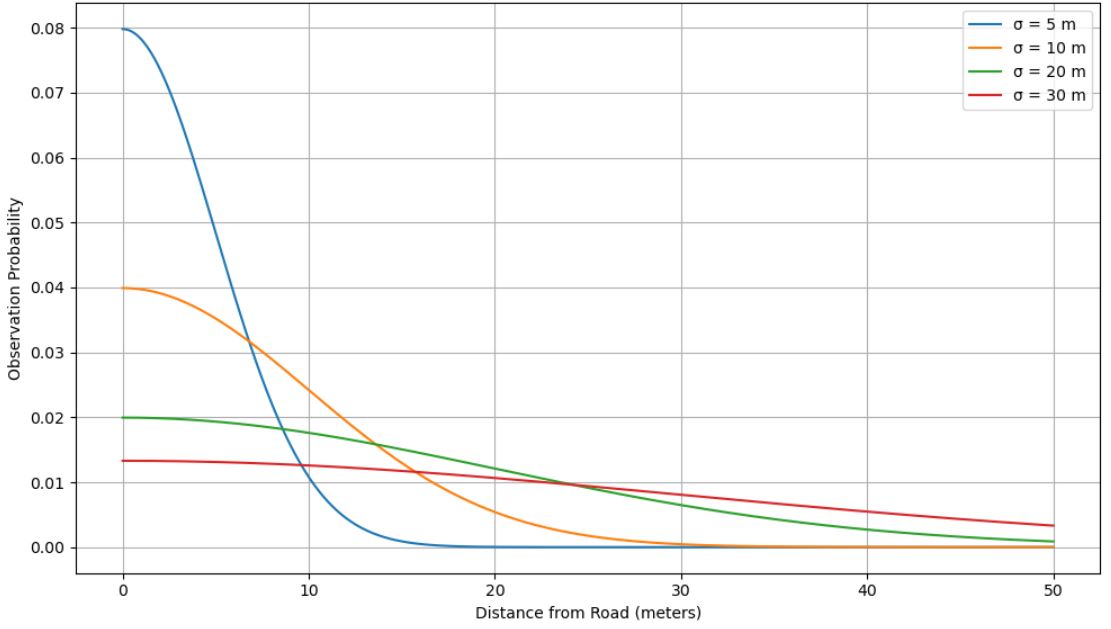}
    \caption{How observation probability changes with distance for different $\sigma$ values. }
    \label{fig:simga change}
\end{figure}

Second, the standard deviation of GPS error distribution $\sigma$ can also be updated according to the GPS uncertainty. As the distance between the sample point $p_i$ and the candidate point $c_i^s$ increases, the value of observation probability decreases. Figure \ref{fig:simga change} indicates how the rate of this decrease changes by choosing different values for $\sigma$. A higher $\sigma$ distributes the likelihood more uniformly across candidate points, reflecting greater positional uncertainty. Conversely, a lower $\sigma$ prioritizes candidates in immediate proximity to the sample, applying a sharper penalty as the distance increases to reflect higher confidence in the GPS measurement.

By leveraging real-time GPS uncertainty directly from the dataset, this approach replaces the heuristic-based selection of a fixed 
$\sigma$ with a data-driven parameterization. This adaptive method ensures that the candidate selection process is physically grounded in the data precision, rather than relying on an arbitrary constant.

\subsubsection{Redesigned Temporal Function}
\label{sec:temporal function}
According to the second key assumption in Section \ref{sec:overview}, the GPS track is supposed to observe the speed limits of the roads in the network. The problem with the original temporal function of the ST-Matching is that cosine similarity only accounts for the direction of the two vectors and not their magnitude. Therefore, it does not really fulfill the assumption. 

The new temporal function suggested in this work consists of three penalty terms: (i) \textit{travel time penalty}; (ii) \textit{speed penalty}; (iii) \textit{speed variation penalty}.

The objective of the travel time penalty is to compare the observed travel time recorded between two consecutive GPS points with the travel time estimated by dividing the shortest-path distance between candidates by the average speed-limit of the roads in the path. The observed travel time $\Delta t_{obs}$ and the estimated travel time based on the shortest path $\Delta t_{est}$ are calculated as follows for each candidate pair $c_{i-1}^t,c_{i}^s$ of consecutive points $p_{i-1}, p_i$:

\begin{gather}
\Delta t_{obs}(i) = t_{i} - t_{i-1}, \\[1em] 
v_{avg}(i) = \frac{\text{dist}_e(p_{i-1}, p_i)}{\Delta t_{obs}(i)}, \label{eq:v average}\\[1em] 
\Delta t_{est}(i, s, t) = \frac{\text{dist}_n(c^t_{i-1}, c_i^s)}{v_{avg}(i)}.
\end{gather}
Ideally, $\Delta t_{est}$ must have a value close to $\Delta t_{obs}$. Larger values of $\Delta t_{est}$ suggest that the calculated path probably highly deviates from the true path time-wise possibly due to unnecessary loops or roundabouts. Similarly, much smaller values could mean that the path is unrealistically short. We propose a travel time penalty function evaluating the similarity between these two numbers with values between 0 to 1. Its definition follows in Equation \ref{eq:penalty tt}. The penalty equals to 1 when the observed and estimated travel time match, and decreases toward 0 as the estimated time deviates the observed time in either direction.

\begin{equation}
\label{eq:penalty tt}
    P_{\text{tt}}(c_{i-1}^t\rightarrow c_i^s) = \exp\left(-\left[\ln\frac{\Delta t_{\text{est}}(i,s,t)}{\Delta t_{\text{obs}}(i)}\right]^2\right).
\end{equation}

The speed penalty is designed to compare the average speed estimated from the trajectory data $v_{\text{avg}}$, with the average speed limit of the road segments in a candidate path $v_{\text{lim}}$, and penalizes only over-speeding when the estimated average speed goes beyond the average speed limit. The average observed speed $v_{\text{avg}}$ is defined in Equation \ref{eq:v average} and the average speed limit $v_{\text{lim}}$ is defined in the following Equation \ref{eq:speed_limit} for each candidate pair $c_{i-1}^t,c_{i}^s$ of consecutive points $p_{i-1}, p_i$:

\begin{equation}
v_{\text{lim}}(i,s,t) = \frac{1}{|E_{i,s,t}|} \sum_{e \in E_{i,s,t}} \text{speed}(e),
\label{eq:speed_limit}
\end{equation}
where $E_{i,s,t}=E_{(i-1,s)\rightarrow (i, t)}$ is the set of edges of the road network included in the shortest path between candidate $c_{i-1}^t$ of $p_{i-1}$ and candidate $c_i^s$ of $p_i$.
Similarly to the travel time penalty function, the speed penalty is defined with an exponential shape  in Equation \ref{eq:speed penalty}. It keeps the value of 1 when the average speed is below the limit,  and drops as it goes beyond the limit.
\begin{equation}
\label{eq:speed penalty}
    P_\text{s}(c_{i-1}^t\rightarrow c_i^s) =
\begin{cases}
1 & \text{if } v_{\text{avg}}(i) \leq v_{\text{lim}}(i,s,t), \\
\exp\left(-\left[\ln\tfrac{v_{\text{avg}}(i)}{v_{\text{lim}}(i,s,t)}\right]^2\right) & \text{otherwise.}
\end{cases}
\end{equation}

Final considerations regard the speed variation penalty. Assuming a moving object on the road network of a city, throughout a trajectory, it passes many road segments and intersections, while prefering to follow relatively direct routes rather than changing the road at every intersection. As a result, the speed limit along a real trajectory should not fluctuate dramatically. The speed variation penalty is introduced to capture this behavior. The penalty is defined based on the ratio of the standard deviation to the mean of speed limits along the candidate path. Considering two candidates, $c_{i-1}^t$ and $c_i^s$, and their candidate path $E_{i,s,t}$, we define the mean and standard deviatiton of the speed limits in the candidate path, and thus the speed variation penalty, as follows:
\begin{gather}
    \mu(i,s,t) = v_{\text{lim}}(i,s,t), \\
    \sigma(i,s,t) = \frac{1}{|E_{i,s,t}|} \sum_{e \in E_{i,s,t}} \left(\text{speed}(e) - \mu(i,s,t)\right)^2, \\
    P_\text{sv}(c_{i-1}^t\rightarrow c_i^s)=\cfrac{1}{1+\sigma(i,s,t)/\mu(i,s,t)}.
\end{gather}

After the calculation of these three components, the new temporal score function is computed as the product of these penalties:
\begin{equation}
\label{eq:F-temporal}
    F'_\text{temporal} = P_\text{tt}\; \cdot \;  P_\text{s}\; \cdot \; P_\text{sv}
\end{equation}

\subsubsection{Behavioral Analysis}
\label{sec:behavioral}
While in the baseline method the link between two candidate points is weighted only by the ST score, behavioral analysis introduces a new function that scores the path between the two candidates based on how much the network edges have been used in the past. The purpose of this analysis is to take into account the historical behavior of road users in the Map Matching procedure, prioritizing high-frequency roads as more probable matches. The approach operates on the premise that historical moving patterns serve as a reliable proxy for current routing preferences. This becomes particularly advantageous in the context of low-frequency GPS data, where the scarcity of spatial points increases path ambiguity; historical priors effectively constrain the search space to the most plausible network trajectories.

The computation of the behavioral score is as follows. Using historical GPS trajectores of a specific period of time, we perform the candidate point selection stage of the algorithm using the dynamic buffer around all the GPS points in the historical dataset. Each candidate point is mapped to a road segment of the network graph. For every instance a candidate point is associated with a given edge, a counter for that edge is incremented by one. As a result, each edge in the network contains a usage score representing how often it has probably been used in the past trajectories.
To deal with the wide range and skewness in edge scores, we apply a logarithmic normalization to scale the values between 0 and 1:
\begin{equation}
    \text{norm edge score} = \cfrac{\ln{(\text{edge score} +1)}}{\ln{(\text{max edge score}+1)}}.
\end{equation}

The road network enriched with the new edge usage score plays a key role in the behavioral analysis component of the Map Matching.
Once the normalized scores are computed, they are integrated into the Map Matching process for new trajectories. For each potential route in the candidate graph, a path usage score is computed as the arithmetic mean of the normalized scores of its constituent edges, as in Equation \ref{eq:behavioral}. This metric serves as a probabilistic weight, quantifying the historical traversal density of each candidate path relative to the prior dataset.

\begin{equation}
   \label{eq:behavioral}
   F_\text{behavioral} (c_{i-1}^t\rightarrow c_i^s) = \frac{1}{|E_{i,s,t}|} \sum\limits_{ e \in E_{i,s,t}} \text{norm edge score}(e),
\end{equation}
where $E_{i,s,t}=E_{(i-1,t)\rightarrow(i,s)}$ is shortest path between candidate $c_{i-1}^t$ of $p_{i-1}$ and candidate $c_i^s$ of $p_i$.

\subsubsection{Final Score Functions}
As a result of all the analyses and modifications discussed so far, we define the score functions of two versions of the algorithm. 
The modified ST-Matching uses a score function combining Equations \ref{eq:spatial score} and \ref{eq:F-temporal}: the buffer and $\sigma$ parameters used to calculate Equation \ref{eq:spatial score} are the dynamic versions, while the temporal function is modified according to Equation \ref{eq:F-temporal}.

\begin{equation}
\label{eq:final modified ST} 
    F_{\text{modified ST}} = F'_{\text{spatial}} \cdot F'_{\text{temporal}}
\end{equation}

Similarly, the final Map Matching function for the case of STB-Matching is defined in a way that adds the behavioral as a third component to the modified ST-Matching, as shown in Equation \ref{eq:STB}.

\begin{equation}
\label{eq:STB}
    F_{\text{STB}} = F'_{\text{spatial}} \cdot F'_{\text{temporal}} \cdot F_{\text{behavioral}}
\end{equation}

\subsection{Evaluation Framework}
\label{sec:metrics}
Traditional map-matching accuracy remains the standard when ground truth is available \cite{experimentalEval2018}. However, in its absence, alternative assessment dimensions become essential. This work establishes a framework that shifts the focus toward proxy metrics, evaluating the reconstructed paths based on their structural integrity and operational efficiency. We adopt a comprehensive and innovative evaluation framework that focuses on the internal consistency, efficiency, and structural characteristics of the matched paths. Our metrics are grouped into four main categories: (i) \textit{efficiency metrics}; (ii) \textit{matching quality metrics}; (iii) \textit{topological metrics}; (iv) \textit{speed metrics}. Table \ref{tab:metrics} reports all the metrics and their mathematical formulation.

\begin{table}[tb]
\centering
 \footnotesize
 \renewcommand{\arraystretch}{1.8}
 \caption{Evaluation metrics proposed in this study. In the formulations, $\mathcal{T}$ is the execution time; $S$ is the set of streets, i.e., grouped edges of the road network; the tilde symbol indicates an optimal variable (e.g., $\tilde{c}_i$ is optimal candidate for the $i$-th GPS point).}
\label{tab:metrics}
\begin{tabular}{lll}
\toprule
    \textbf{Category} & \textbf{Name} & \textbf{Formula} \\
\midrule
    Efficiency & Runtime [seconds] & $E_1(T) = \mathcal{T}_{end} - \mathcal{T}_{start}$ \\
    & Average number of candidate points & $E_2(T) = \frac{1}{n}\sum m_i$ \\
    & Total number of candidate points & $E_3(T) = \sum m_i$\\
\midrule
    Quality & Average projection distance [meters] & $Q_1(T) = \frac{1}{n} \sum \text{dist}_e(p_i, \tilde{c}_i)$ \\
    & Length metric & $Q_2(T) = \frac{\sum\text{dist}_e(p_i, p_{i-1})}{\sum\text{dist}_n(\tilde{c}_i,\tilde{c}_{i-1})}$\\
    & Complexity ratio & $Q_3(T) = \frac{\sum\text{dist}_n(\tilde{c}_i,\tilde{c}_{i-1})}{\text{dist}_e(p_1,p_n)}$ \\
\midrule
    Topology & Number of revisited edges & $T_1(T) = |\{\tilde{e}\in E: \text{count}(e)>1\}|$ \\
    & Number of revisited streets & $T_2(T)=|\{\tilde{s}\in S: \text{count}(s)>1\}|$ \\
    & Number of loops & $T_3(T) = |\{\text{loops}\in \tilde{P}\}|$ \\
\midrule
    Speed & Speed relative deviation & $ S_1(T) = \frac{|\text{Sample Speed }-\text{Path Speed}|}{\text{Sample Speed}}$\\
               
\bottomrule
\end{tabular}
\end{table}

\subsubsection{Efficiency Metrics}
The efficiency metrics assess the computational efficiency of the algorithm. By evaluating these metrics, we quantify the overhead introduced by different candidate selection strategies (e.g., fixed vs. dynamic buffers) and identify trade-offs between algorithmic complexity and matching accuracy. Specifically, we look at the following quantities per each trajectory $T$ of the dataset:
\begin{itemize}
    \item {Runtime:} The total time it takes for the algorithm to run on a trajectory. It indicates whether the algorithm is appropriate for deployment in production settings and whether it can handle high-frequency data streams without introducing latency.
    \item {Average number of candidate points:} This serves as a proxy for the complexity of the shortest-path search. A high average indicates a dense search space, which may lead to increased memory consumption, whereas a lower average suggests a more pruned and efficient candidate selection.
    \item {Total number of candidate points:} This cumulative measure allows us to understand how the algorithm scales with the trajectory length and the geometric complexity of the underlying road network.
\end{itemize}

\subsubsection{Matching Quality Metrics}
The matching quality metrics act as a proxy to assess the quality of the matched path in absence of the current ground truth. They are defined as:
\begin{itemize}
    \item {Average projection distance:} The average distance between each GPS point and its corresponding matched candidate point on the road network. When comparing two different algorithms, this metric can help understand if candidate selection has changed due to a different buffer or the effect of temporal or behavioral scoring.
    \item {Length metric:} This ratio measures the geometric fidelity of the matched path relative to the observed GPS sequence. Since GPS samples typically follow the road geometry, a ratio significantly different from 1 may indicate ``over-matching" (i.e., the path is too long due to unnecessary loops or zig-zags) or ``under-matching" (i.e., the path is too short, potentially skipping road segments), indicating a loss of geometric fidelity.
    \item {Complexity ratio:} The ratio of the matched path length to the Euclidean distance between the first and last GPS points of a trajectory. The metric checks how much the matched path deviates from a direct line between the origin and the destination, quantifying the tortuosity of the reconstructed route.
\end{itemize}

\subsubsection{Topology Metrics}
The topology metrics are designed to capture the topological properties of the matched path and measure potential inconsistencies or redundancies. Higher values in these metrics may indicate unrealistic paths or abnormal behavior.

\begin{itemize}
    \item {Number of revisited edges:} The number of times the same edge in the road network graph is used more than once for the same trajectory. Frequent revisits are typically physically implausible for standard transit and suggest that the algorithm is ``trapped" oscillating between candidate edges.
    \item {Number of revisited streets:} Beyond individual road edges, this metric quantifies the number of times that different segments of the same street are used more than once. This is a key indicator of detour errors, where the algorithm might have matched a point to a side street or a parallel service road before returning to the main thoroughfare.
    \item {Number of loops:} Defined by the number of polygons formed by the matched path. While naturally present in roundabouts, an excess of loops identifies topological instability and physically-implausible tracing caused by high GPS noise or weak temporal constraints.
\end{itemize}

\subsubsection{Speed-Based Metric}
The speed-based metric evaluates how realistic the reconstructed travel speed is compared to the expected speed estimated from the GPS samples. 
The speed estimated by the GPS data is treated as a reference. The metric is calculated as the speed relative deviation.

\section{Case Study and Results}
\label{sec:case study}

\subsection{GPS Trajectory Dataset}

The GPS trajectory dataset used in this work is provided by Cuebiq Srl, a company specializing in location intelligence. The data is collected through a GDPR-compliant opt-in process from users who have exclusively consented to anonymous data sharing. The dataset contains anonymous GPS tracks collected from mobile devices. Each record includes geographical coordinates and timestamps, as well as additional features such as trajectory ID and GPS uncertainty, as described in Table \ref{table:gps_features}.

\begin{table}[tb]
    \caption{Main features of the GPS trajectory dataset.}
    \label{table:gps_features}
    \footnotesize
    \centering 
    \begin{tabular}{lll} 
    \toprule
    \textbf{Field} & \textbf{Type} & \textbf{Description} \\
    \midrule
    {Trajectory ID} & categorical & Unique identifier for each individual trajectory \\
    {Latitude} & numerical & Latitude coordinate of the GPS observation \\
    {Longitude} & numerical & Longitude coordinate of the GPS observation \\
    {Uncertainty} & numerical & Radius of a circle around the GPS point with 68\% probability \\
    {Timestamp} & numerical & Date and time at which the GPS point was recorded \\
    \bottomrule
    \end{tabular}
\end{table}

The trajectory observations analyzed in this study are located in the urban area of Milan, Italy, and its surroundings and refer to the period from December 1st to December 20th, 2019. 
The initial dataset is subjected to a filtering process to guarantee the high quality of the location data and its consistency with the objectives of the study. While the focus of the analysis is on the urban area of Milan, the dataset contains many records outside this area, probably related to daily commuters. Therefore, the first filter is to exclude the records outside the urban area, which is done using the official shapefile of the Milan urban area from the open data portal of the municipality \cite{milano-opendata}. 

The sampling interval of the GPS records is quite high in most cases, with the average time interval of 16 seconds over the whole dataset, which means that this is a case of high-frequency GPS data. In order to make sure that the trajectories that are potentially going to be used later in Map Matching are long enough, only the ones with at least 10 GPS records throughout the trajectory are kept. 

Finally, in this study, the focus is on the vehicular mobility on the drivable road network, which means trajectories corresponding to walking users must preferably be disregarded. This is achieved by calculating an approximate average speed for each trajectory using the direct distance between each two adjacent points and the time interval between them, and keeping the trajectories with an average speed of at least 6km/h to make sure there are no walking users in the dataset. In the end, the final dataset is composed of 
trajectory IDs corresponding to motor vehicles moving in the city of Milan. 

\subsection{Road Network Description}
\label{sec:road network}

The road network used in this paper covers the urban area of Milan. It is retrieved from OpenStreetMap \cite{OpenStreetMap} using the \texttt{OSMnx} Python library \cite{boeing2025modeling}. The network type \texttt{drive} is used in this case, which includes only drivable roads. The retrieved graph is directed and multi-edge, meaning multiple edges may exist between a pair of nodes. 

Initially downloaded in the WGS 84 coordinate reference system (EPSG:4326), the network is converted into a projected coordinate reference system, ETRS89 / UTM Zone 32N (EPSG:6707), to enable precise spatial computations in meters, such as path lengths and distances between GPS observations.

The road network contains $12891$ nodes and $25404$ edges. The graph has an average node degree of 3.94, with an average in-degree and out-degree of 1.97, indicating a balanced directionality. The network has a diameter of 169 and a relatively low density of approximately 0.00015. This density is a graph-theoretic concept, and having a low value is the characteristic of urban graphs, since in the street network of a city, there are typically many nodes and each node is only connected to a few nodes around itself. However, the network in the spatial terms, is practically dense. 

Each edge in the network contains a range of attributes such as: \textit{start node}, \textit{end node}, \textit{OSM ID}, \textit{street name}, \textit{highway type}, \textit{length}, \textit{geometry}, and \textit{maximum speed} of the road. The average edge length in the network is approximately 98.5 meters, having a strongly right-skewed distribution, with the majority of road segments measuring less than 100 meters. This is also normally expected from road networks of the cities and shows that nodes are mostly close to each other, resulting in a spatially high-density network. Among road types, the most common in the dataset includes \textit{motorway, truck, primary, secondary, tertiary, residential} and \textit{unclassified.}

The speed attribute of the graph edges showed missing or non-numerical values, a common limitation in the OpenStreetMap database. Since this attribute is essential for the temporal analysis of Map Matching, the missing value imputation is addressed in several steps. First, in cases where there are no numerical values, but instead a road type code, it is possible to infer information about the speed limits for the corresponding code from the OpenStreetMap Wiki \cite{wiki:maxspeed}. In few cases where there is a list of two values for the speed limit, the greater values are kept. Later, dealing with missing values, assuming that different segments of the same street usually have the same speed limit, we use the \texttt{name} attribute (i.e., the street name) to fill in the missing values if other segments of the street with the same name contain a maximum speed value. Finally, for the rest of the missing values, another attribute called \texttt{highway} (i.e., the highway type) is used considering the mode of maximum speed values for each road type. The final distribution of the speed limits of the network edges comprises a majority of road segments of 30 km/h (56.0\%), followrd by many road segments of 50km/h (38.1\%), and the remaining with other velocities.

\subsection{Results}
\label{sec:results}
This section presents the results of the Map Matching using different methods and compares them through the evaluation metrics discussed in Section \ref{sec:metrics}.

For the purpose of running the algorithm and performing analysis, a random sample of 500 trajectories is taken. Then, two Map Matching algorithms are carried out: the traditional ST-Matching, and the modified ST-Matching with dynamic buffer, dynamic observation probability, and redesigned temporal function. In case of original ST-Matching, the fixed buffer is fixed to 50 meters, which according to the network characteristics discussed in Section \ref{sec:road network} is large enough to contain at least a few candidate points per GPS sample. The parameter $\sigma$ in the observation probability is set to 20 meters. For the modified ST-Matching, whenever the value of GPS uncertainty is higher than 50, the value of 50 is used as the maximum buffer. 

To simulate a low-frequency trajectory, a subsample of 100 trajectories is considered and limited: for each trajectory from the first GPS point we keep the poits with at least a 2 minute time interval from the previous one. The low-frequency trajectories are then used to compare ST-Matching with the STB-Matching version including behavioral analysis. The evaluation is performed exploiting the proposed metrics, but also comparing the results obtained with the synthetic low-frequency data with the ones of their original counterpart, which is used as a ground truth. Moreover, for the STB-Matching, the edge score of the road network is computed using a different subsample of size 400 of the trajectory dataset.

\subsubsection{ST-Matching vs. Modified ST-Matching}
This section compares the result of Map Matching between the original ST-Matching and the modified version which includes dynamic buffer, dynamic observation probability, and the new temporal function. The comparison is carried out through the metrics mentioned in Section \ref{sec:metrics}. It is based on the distribution of the metric values and the statistical significance expressed by a t-test.

\begin{figure}[tb]
    \centering
    \includegraphics[width=0.75\linewidth]{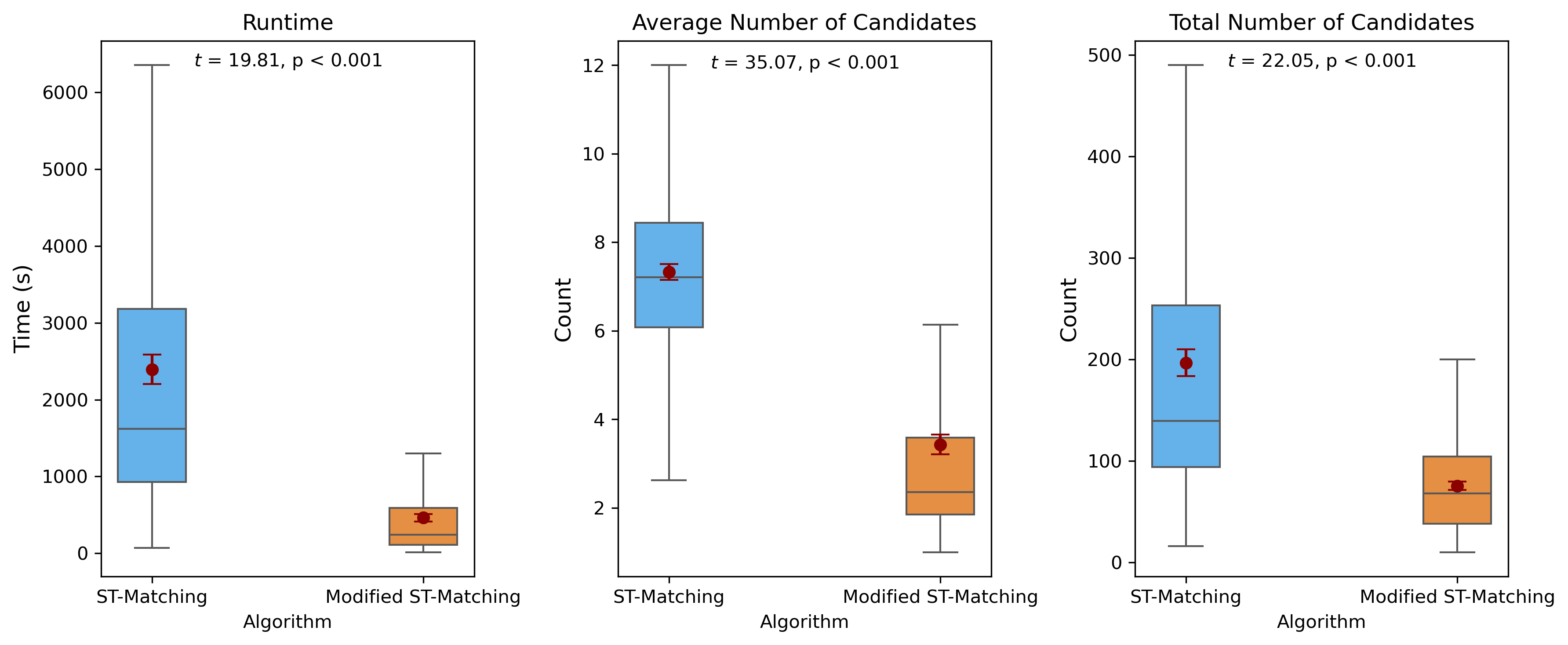}
    \caption{ST-Matching vs. Modified ST-Matching across efficiency metrics.}
    \label{fig:efficiency metrics hf}
\end{figure}

The results of the analysis clearly show that there is a significant improvement in the efficiency metrics. As a result of using a dynamic buffer, in cases of points with better accuracy, the search area and therefore the number of candidate points have decreased, which leads to much lower runtime. Figure \ref{fig:efficiency metrics hf}, which illustrates the distribution of the metric values through boxplots along with the result of the t-test, confirms this significance. 

\begin{figure}[tb]
    \centering
    \includegraphics[width=0.75\linewidth]{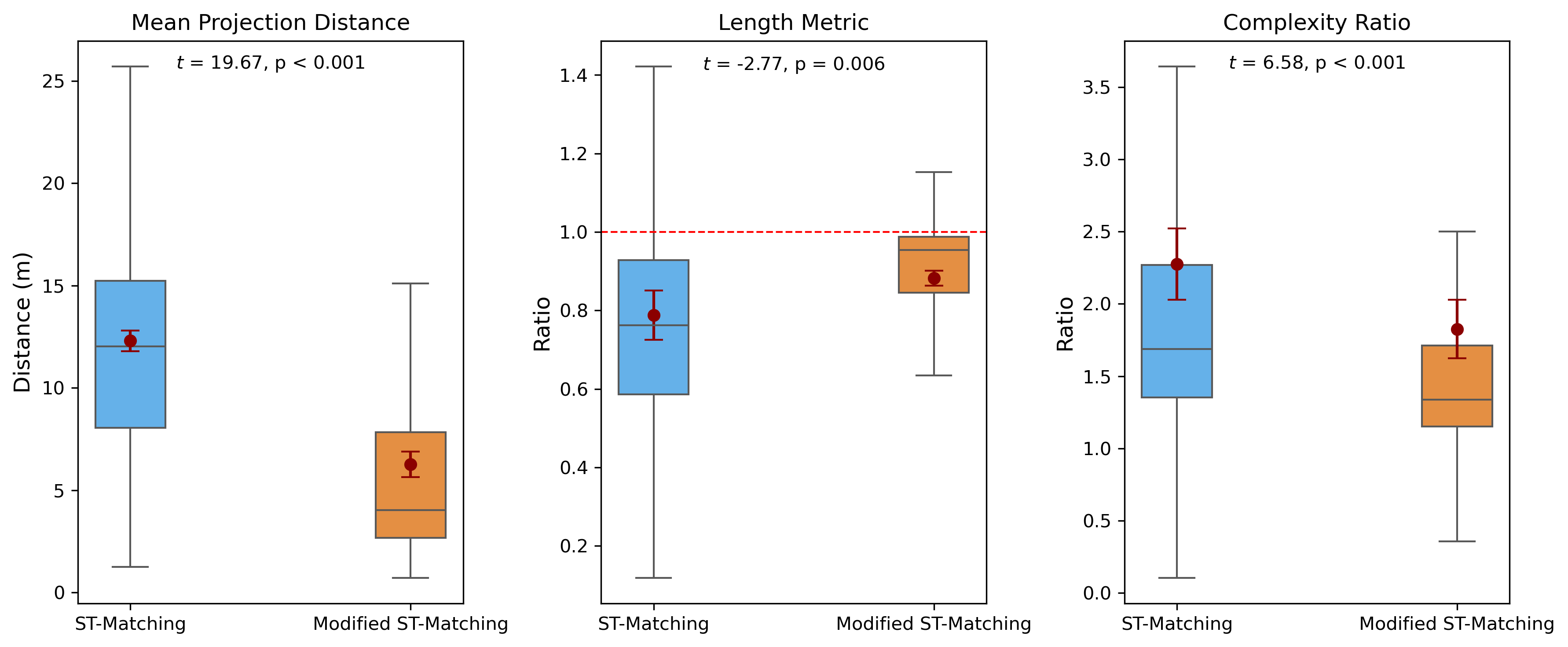}
    \caption{ST-Matching vs. Modified ST-Matching across matching quality metrics.}
    \label{fig:Quality metrics hf}
\end{figure}

As can be seen in Figure \ref{fig:Quality metrics hf}, the change in matching quality metrics is also statistically significant. Mean projection distance is reduced in the modified ST-Matching. This reflects the fact that the algorithm tends to choose candidate points that are spatially closer to the sample point, potentially due to tighter buffers and dynamic observation probability. Although this metric gives useful insight about how the algorithm chooses the candidate points, it cannot evaluate the performance of the algorithm alone and must be considered in conjunction with other metrics. Values of the length metric in the modified algorithm are more distributed around 1, which means the paths are probably more realistic. Complexity ratio also experiences a reduction, implying that the matching paths in the modified algorithm are less fragmented and generally simpler, in other words, closer to a straight line.

\begin{figure}[tb]
    \centering
    \includegraphics[width=0.75\linewidth]{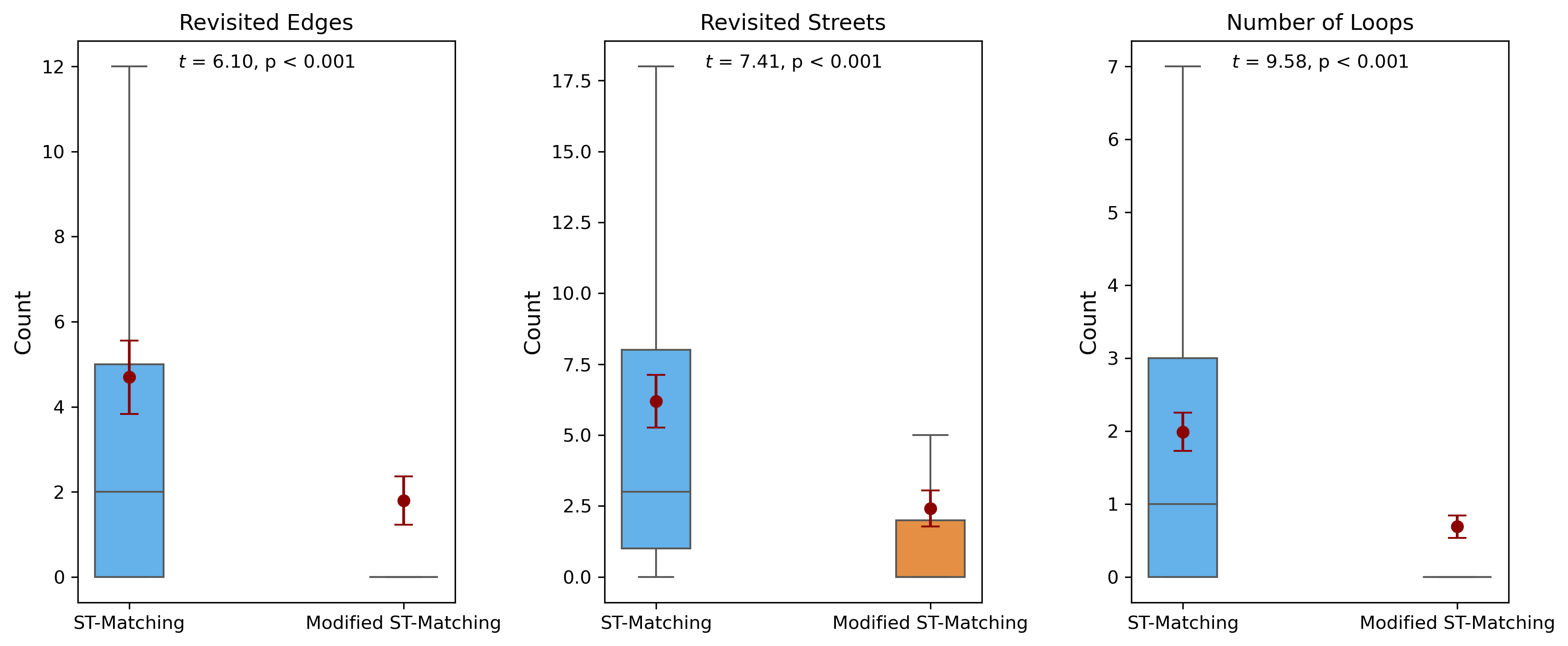}
    \caption{ST-Matching vs. Modified ST-Matching across topology metrics.}
    \label{fig:topology metrics hf}
\end{figure}

Topology metrics are also improved. The boxplots in Figure \ref{fig:topology metrics hf} illustrate that revisited network edges and the number of loops in the paths are almost removed. This may possibly be the effect of a more realistic temporal function (see the supplementary analysis in \cite{github-repo-ali}). Fewer revisited streets can be interpreted as fewer indirect paths. Thus, we can claim that the modified algorithm fulfills the basic assumptions of the original ST-Matching algorithm better.

\begin{figure}[tb]
    \centering
    \includegraphics[width=0.28\linewidth]{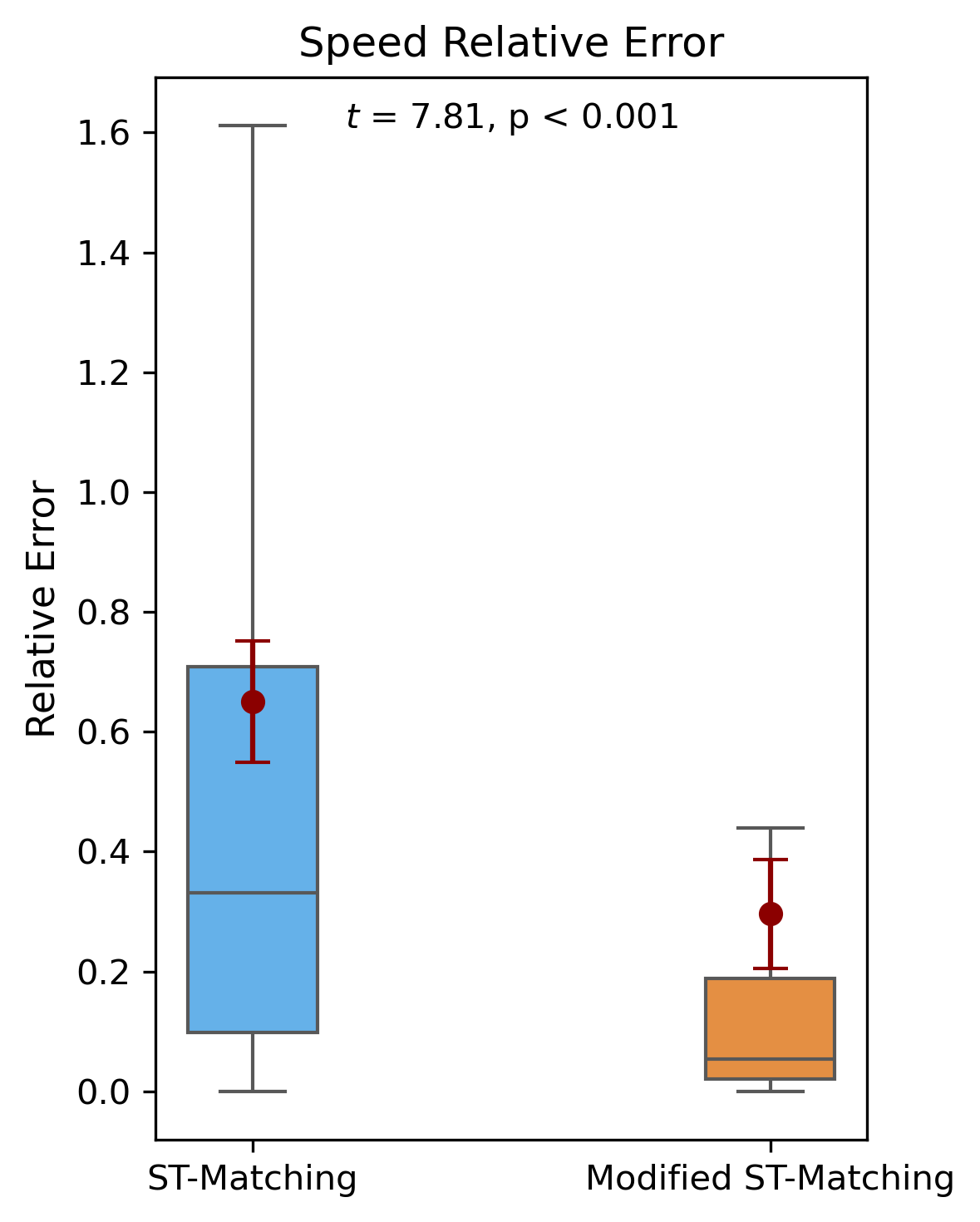}
    \caption{Comparison of ST-Matching and Modified ST-Matching based on the speed metric. }
    \label{fig:speed metric hf}
\end{figure}

It is observed in Figure \ref{fig:speed metric hf} that the speed relative deviation, similar to other metrics, signals an improvement in the results and potentially more realistic paths regarding the temporal aspects of the Map Matching process.

\begin{figure}[tb]
    \centering
    \includegraphics[width=0.9\linewidth]{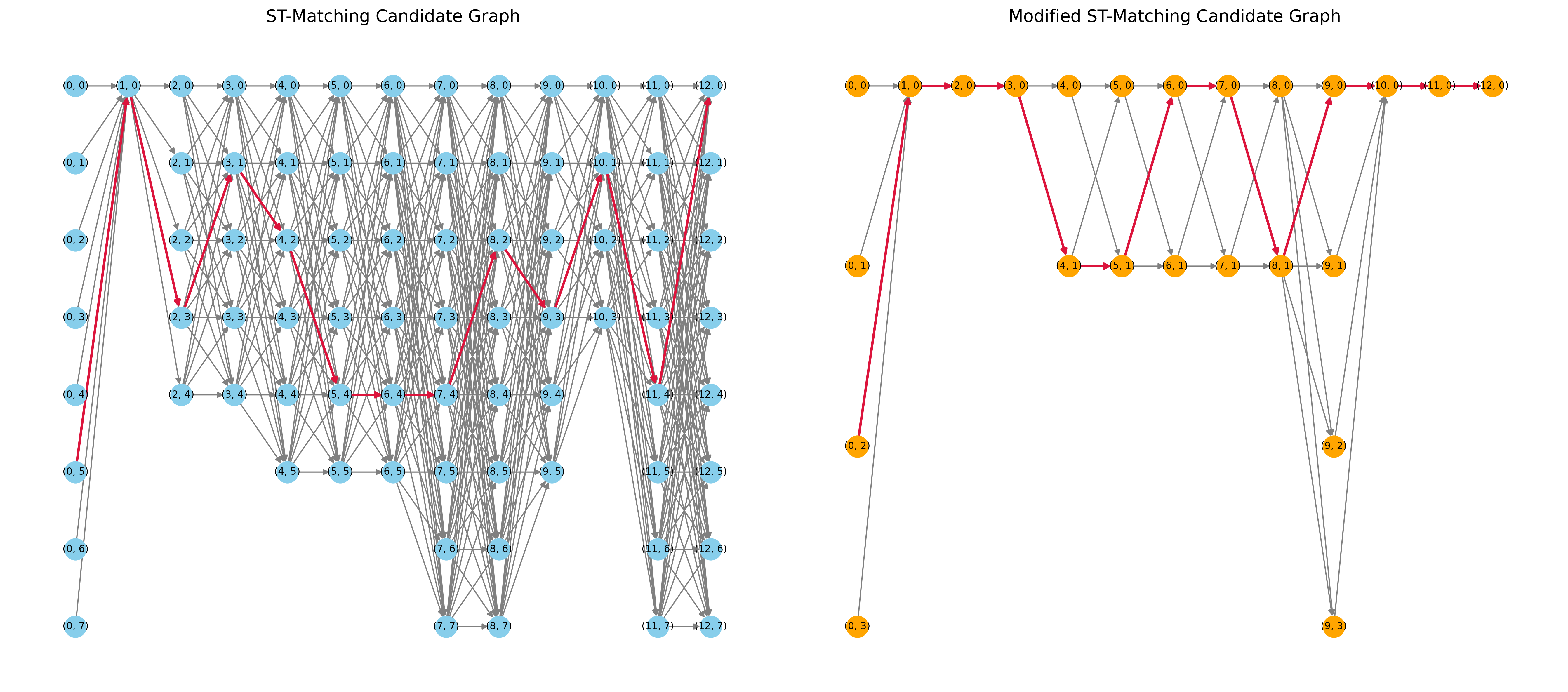}
    \caption{Comparison of the size of candidate graphs between ST-Matching and Modified ST-Matching (The red line represents the selected matched path).}
    \label{fig:candidate graphs}
\end{figure}

\begin{figure}[tb]
    \centering
    \begin{minipage}[t]{0.48\textwidth}
        \centering
        \includegraphics[width=\linewidth]{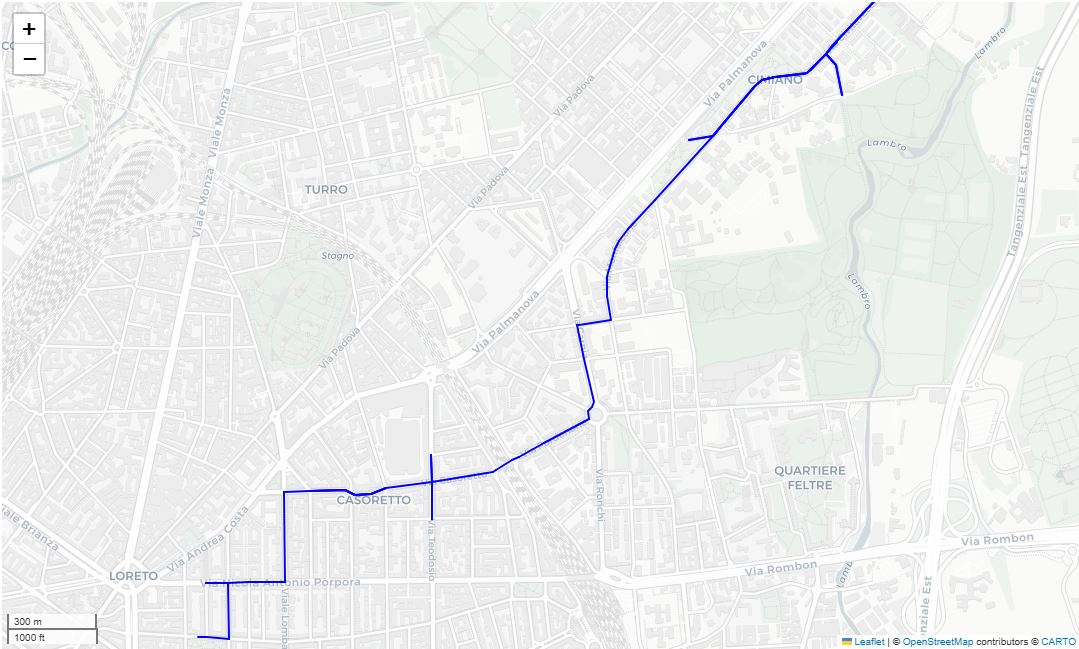}
        \caption*{ST-Matching}  
    \end{minipage}
    \hfill
    \begin{minipage}[t]{0.48\textwidth}
        \centering
        \includegraphics[width=\linewidth]{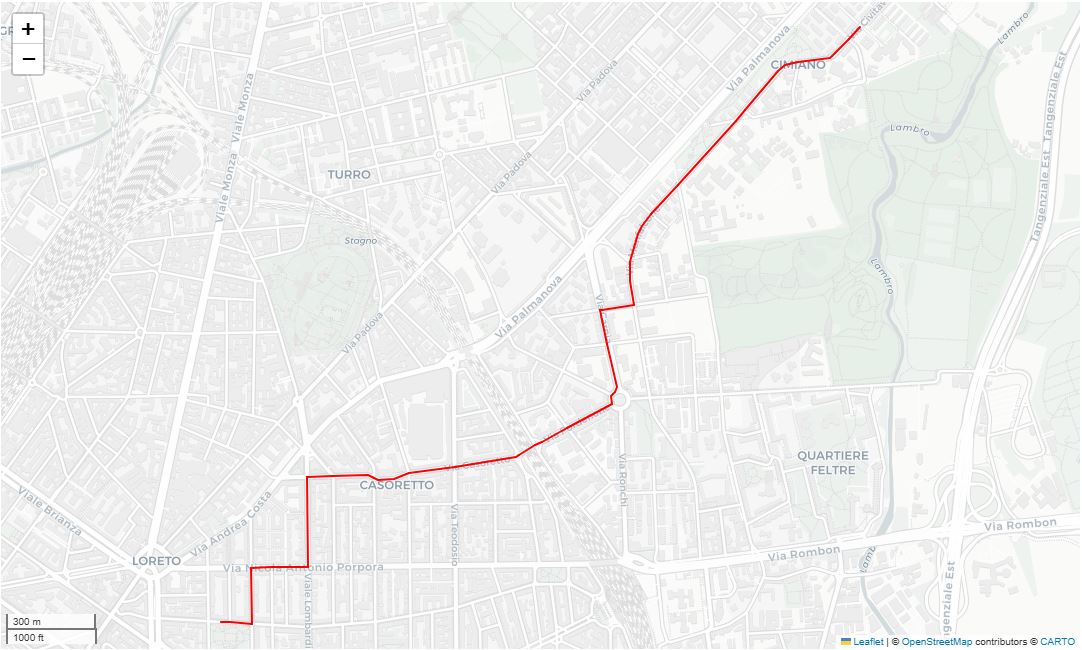}
        \caption*{Modified ST-Matching}  
    \end{minipage}
    \caption{An example of a reconstructed path for the same trajectory using the two different algorithms.}
    \label{fig:side_by_side_sample}
\end{figure}

Overall, the modified ST-Matching algorithm creates noticeable positive changes across all evaluation metrics. Since the 50 meter buffer is potentially too large for a dense urban network like Milan, and the value of GPS accuracy in many samples is lower than 50 (lower uncertainty), the search buffer has been much smaller on average across many trajectories, resulting less candidate points and therefore smaller candidate graphs. This is very well reflected in efficiency metrics. Figure \ref{fig:candidate graphs} presents an example of a trajectory's candidate graphs created by ST-Matching and its modified version, and the remarkable reduction in the size of the candidate graph in the case of the modified ST-Matching. Figure \ref{fig:side_by_side_sample} shows an example of the path created by the two algorithms using the same trajectory. It can be seen how the modified algorithm avoids sharp detours that lead to revisited streets.

\subsubsection{ST-Matching vs. STB-Matching}
The same comparison across all metrics is presented in this section using to evaluate the original method and the STB variant. 

\begin{figure}[tb]
    \centering
    \includegraphics[width=0.5\linewidth]{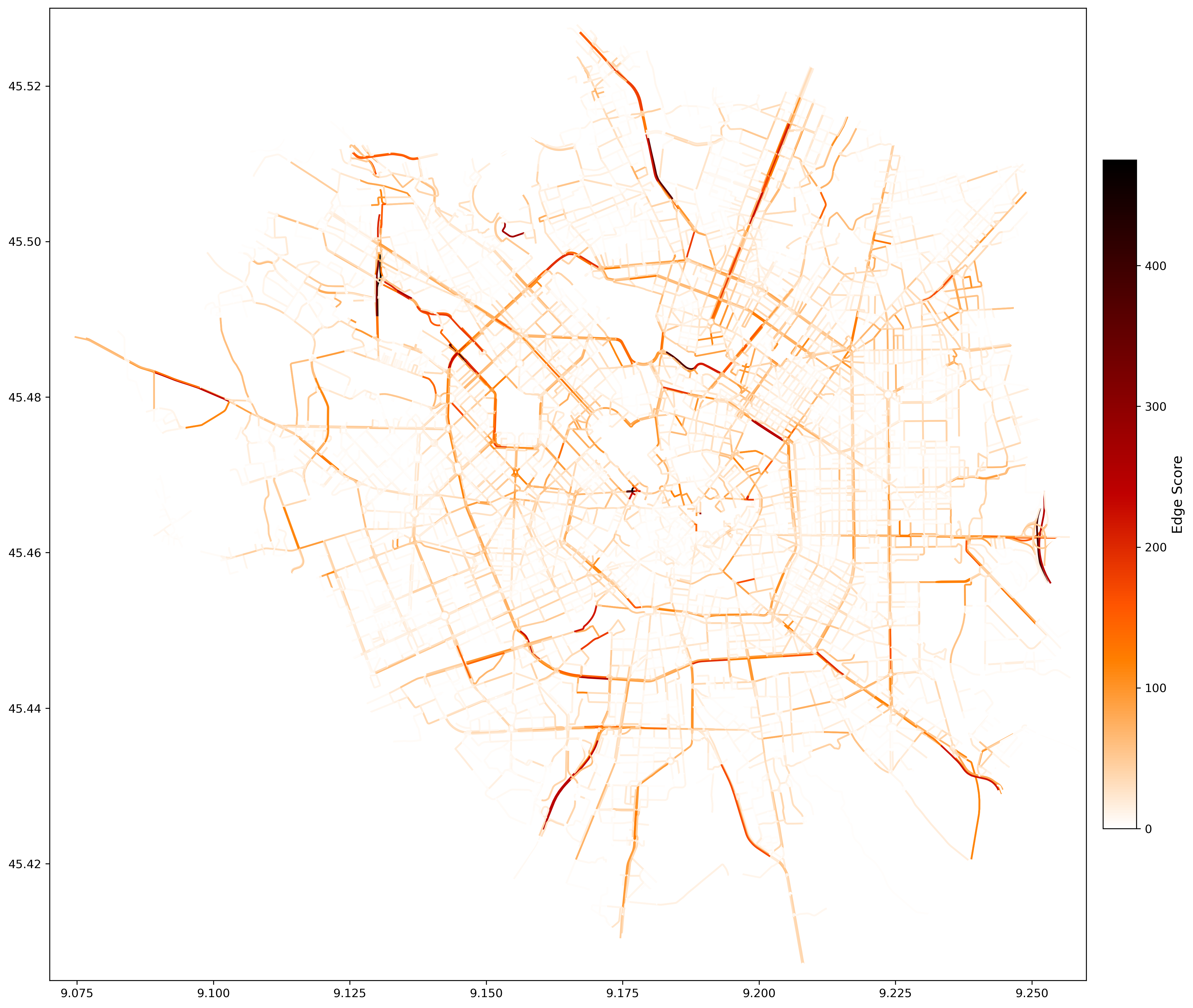}
    \caption{Edge scores of the road network of Milan.}
    \label{fig:edge score}
\end{figure}

First, the edge score of the behavioral analysis is computed with 400 historical trajectories. Figure \ref{fig:edge score} illustrates the distribution of the edge score on the road network of the urban area of Milan using the available dataset. It can be observed that the roads directed towards the suburban areas and the central rings are generally darker, meaning that they have higher scores and have been used more frequently according to the available dataset.

\begin{figure}[tb]
    \centering
    \includegraphics[width=0.75\linewidth]{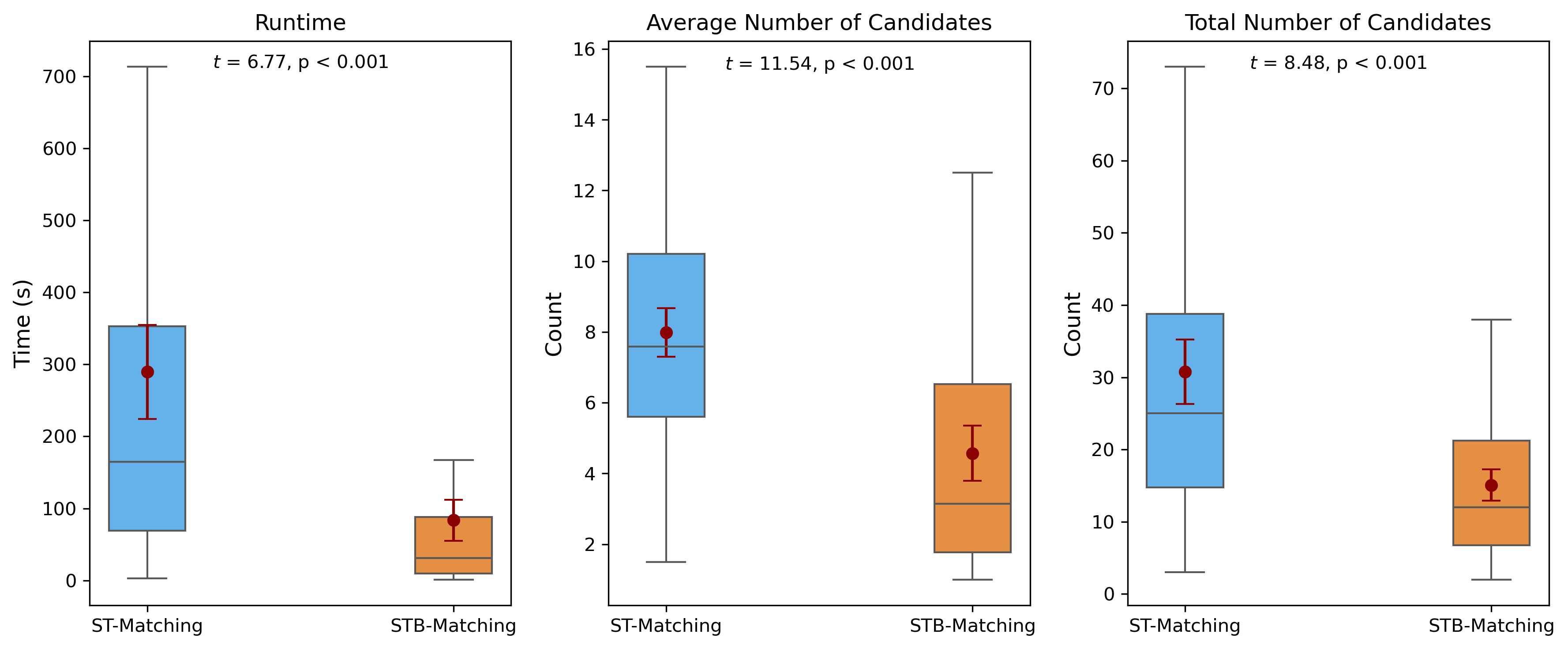}
    \caption{ST-Matching vs. STB-Matching across efficiency metrics.}
    \label{fig:efficiency metrics lf}
\end{figure}

For a comprehensive comparison, we use the lower-frequency synthetic sample of 100 trajectories of the dataset to assess the main metrics, and their original counterpart as a ground truth. 
As expected, efficiency metrics are positively affected by the new algorithm due to the dynamic buffer that results in less candidate points and less runtime. The result of the t-test, as written on the graphs in Figure \ref{fig:efficiency metrics lf}, confirms this effect.

\begin{figure}[tb]
    \centering
    \includegraphics[width=0.75\linewidth]{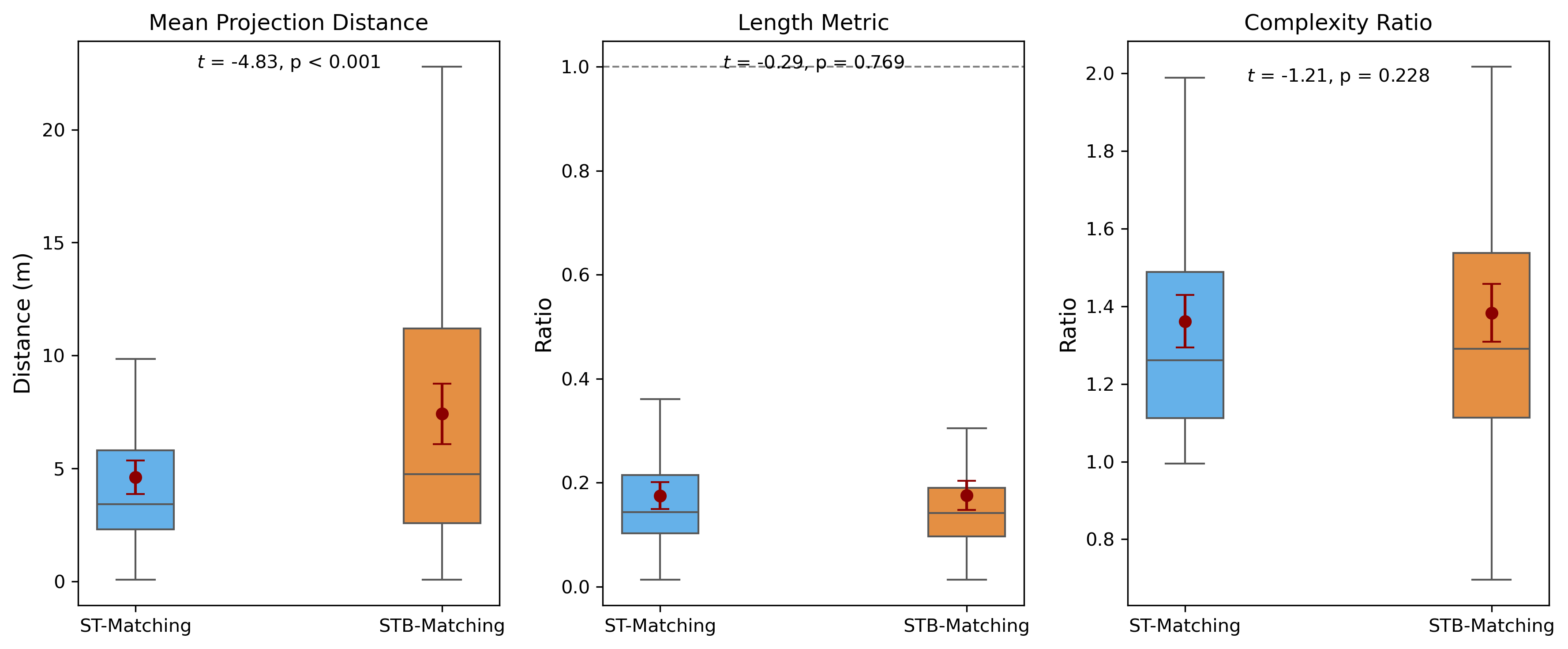}
    \caption{ST-Matching vs. STB-Matching across matching quality metrics.}
    \label{fig:quality metrics lf}
\end{figure}

Unlike the case of high frequency data, length metric and complexity ratio do not show any major change in the distribution of their values (Figure \ref{fig:quality metrics lf}). However, the change in mean projection distance is statistically significant and its range has increased. This tells us that the STB-Matching chooses the candidate points that are on average spatially further than the GPS sample.

\begin{figure}[tb]
    \centering
    \includegraphics[width=0.75\linewidth]{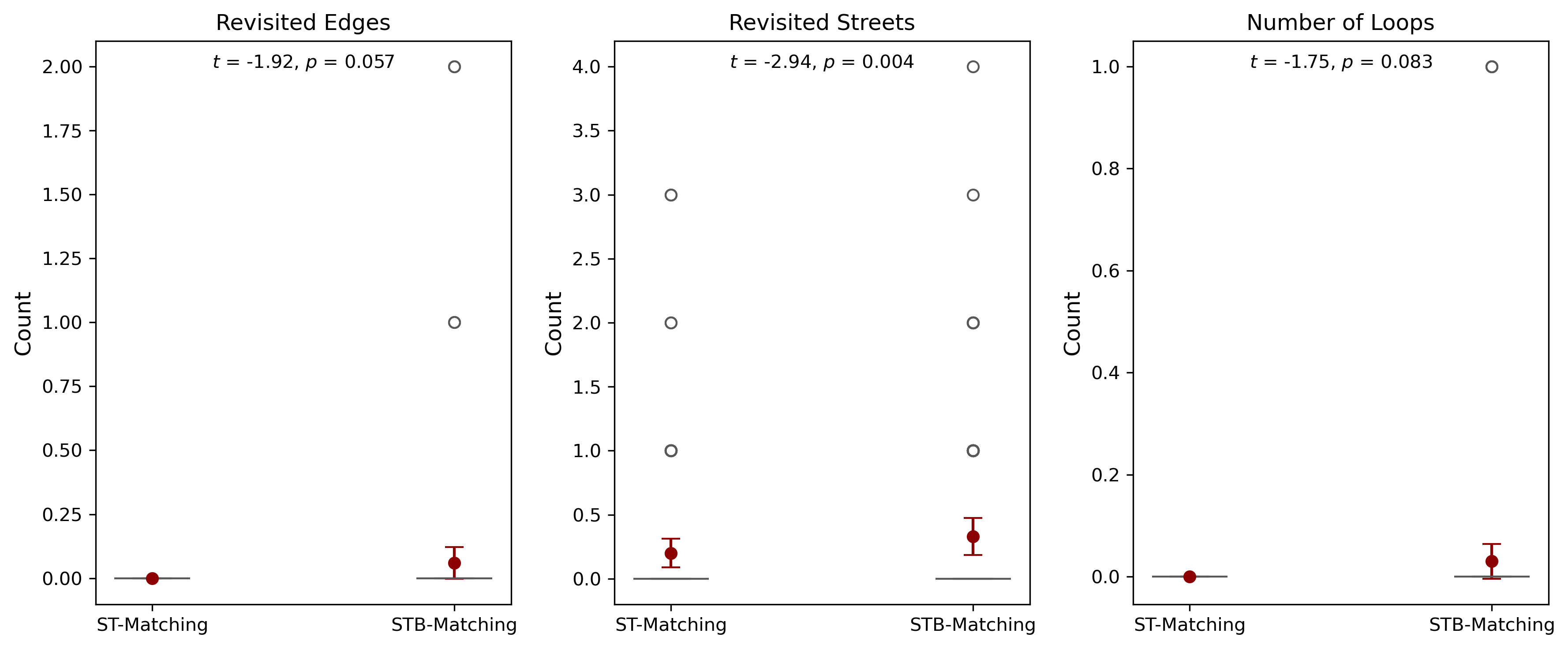}
    \caption{ST-Matching vs. STB-Matching across topology metrics.}
    \label{fig:topology metrics lf}
\end{figure}

Regarding the topology metrics, Figure \ref{fig:topology metrics lf} shows that in almost all trajectories, the value of these metrics is zero. There are only few cases of non-zero values. Therefore, both algorithms handle the revisited streets and edges and also the path loops well. 
In a trajectory with low frequency data, the distance between neighboring points is usually higher. The A* algorithm \cite{hart1968formal}, which searches for the shortest path between such points, typically avoids going through loops or the same streets because then it will not be the shortest path. The reason we see more loops and revisited streets and edges in the case of high frequency data is that more sampling points can also increase the error. One wrongly matched candidate point between two correctly matched ones can deviate the final matched path from the real path and cause unnecessary loops in order to come back on the right track.

\begin{figure}[tb]
    \centering
    \includegraphics[width=0.28\linewidth]{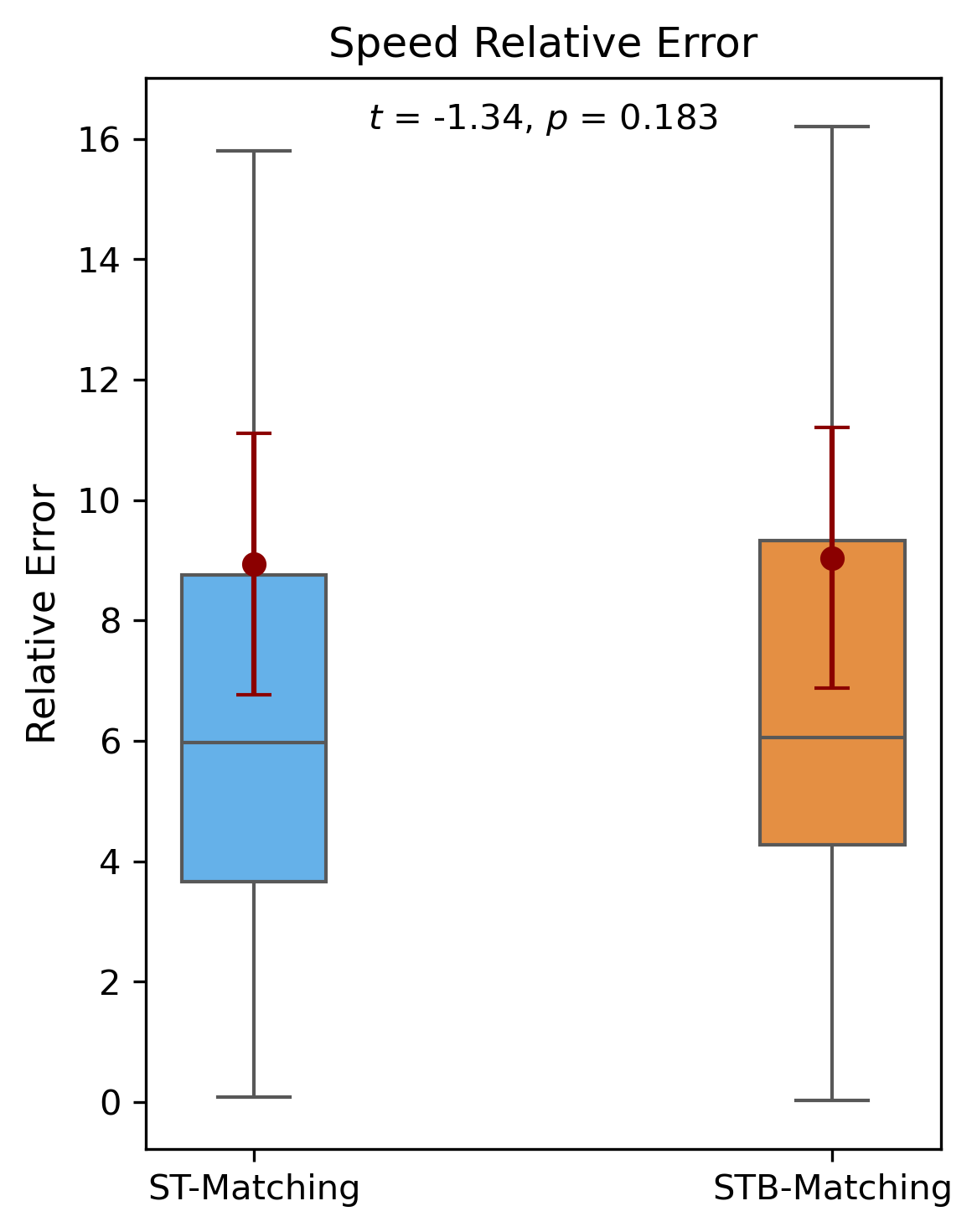}
    \caption{Comparison of ST-Matching vs. STB-Matching based on speed metric.}
    \label{fig:speed metric lf}
\end{figure}

The speed relative deviation also does not show any significant differences between the two algorithms. It is clearly observed in Figure \ref{fig:speed metric lf} that the distribution of the values and the t-test results indicate almost no change in this metric.

Finally, we compare the results of the path enhancement using the synthetic low-frequency data and their original counterpart, assuming the matched path of the latter as the ground truth and checking how close are the results of ST-Matching and STB-Matching using the former. In order to do so, the edges of the final matched paths by ST-Matching and STB-Matching are compared with that of the modified ST-Matching to see the number of edges they have in common. As summarized in Table \ref{table:edge_overlap_percentage}, among 100 trajectories, in 72 cases ST-Matching and STB-Matching had the same number of edges in common with the modified ST-Matching results. STB-Matching demonstrated superior performance in 15\% of instances, while ST-Matching proved more effective in 13\% of cases. This reveals that improving the performance of ST-Matching algorithm for low frequency GPS data is a demanding job and might need more investigation.

\begin{table}[t]
    \footnotesize
    \centering 
    \caption{Overlap between ST-Matching and STB-Matching using synthetic low-frequency data with the results of the modified ST-Matching on original high-frequency data.}
    \label{table:edge_overlap_percentage}
    \begin{tabular}{lc}
    \toprule
    \textbf{Comparison Outcome} & \textbf{Percentage} \\
    \midrule
    Equal overlap with modified ST-Matching & 72\% \\
    STB-Matching closer to modified ST-Matching & 15\% \\
    ST-Matching closer to modified ST-Matching & 13\% \\
    \bottomrule
    \end{tabular}
\end{table}

\section{Conclusions and Discussion}
\label{sec:conclusion}
This paper aimed at enhancing a {Map Matching} algorithm for low frequency data, known as the {ST-Matching}, both to improve its computational efficiency and the precision of the results. The study explored several modifications by taking into account the available features of the data, such as {GPS uncertainty} and the historical usage patterns of the road network.
To achieve this goal, four major modifications were suggested: a \textit{dynamic buffer} for candidate point search based on GPS uncertainty; a \textit{dynamic observation probability} function where its parameters change with respect to GPS uncertainty; an entirely \textit{redesigned temporal score} consists of three penalty functions; and a \textit{behavioral function} based on historical data. They led to the implementation of two new enhanced version of the ST-Matching algorithm: the \textit{modified ST-Matching}, and the \textit{STB-Matching}.

Given the absence of ground truth data, the proposed Map Matching enhancements were evaluated through a novel assessment framework. This original methodology employs a multidimensional set of internal metrics, specifically categorized into \textit{efficiency}, \textit{quality}, \textit{topology}, and \textit{speed} metrics, thereby providing a robust and comprehensive validation of the algorithm's performance across diverse operational constraints.

The algorithms were compared using real-world trajectory data located in the municipality of Milan, Italy. The analyses were carried out both on the original high-frequency data and the synthetic low-frequency data generated from the original ones by removing observations. 
The results demonstrated that the dynamic buffer significantly reduces computational runtime by limiting the search area in case the dataset has a relatively high GPS accuracy. This alone also impacts the accuracy of the results since choosing the correct candidate point is a vital part of the algorithm in order to guess the correct matching path. The redesigned temporal function makes the algorithm topologically more consistent by removing the loops and revisited edges or, in general, the unrealistic segments of the path. Overall, the {modified ST-Matching} algorithm achieved consistently better results in comparison with the baseline across all evaluation metrics. However, when it comes to low frequency GPS data, although the improvement in the efficiency of {STB-Matching} is still remarkable, the final output of the algorithm barely changes. It can be inferred that Map Matching in the realm of low frequency GPS data is still a challenging task which requires probably more sophisticated scoring models or richer contextual information. 

Several research can be identified to extend the findings of this study. Primarily, obtaining more granular insights into the dataset's metadata, specifically regarding the {nature of the trips} and the {vehicle classifications}, would significantly enhance the behavioral analysis introduced herein. For instance, if edge scores were derived from a longitudinal dataset of daily commuters, these metrics would inherently provide higher fidelity for the Map Matching of trajectories belonging to the same user profile. In the present study, the absence of such categorical information necessitated a more generalized approach.

Furthermore, the development of more sophisticated {penalty functions} within the temporal scoring framework represents a critical area for future investigation, aimed at improving the precision and robustness of the scoring mechanism. It is important to note that this research focused exclusively on the modification of score functions, specifically, the calibration of weights for the nodes and links within the candidate graph, while the core path-finding logic remained reliant on the standard A* algorithm \cite{hart1968formal}.

Future work should consider structural modifications to the A* algorithm itself or the implementation of alternative shortest-path heuristics. While Map Matching performance is typically sensitive to the selection of candidate points, the challenges posed by low-frequency data in {STB-Matching} mean that even with optimal candidate selection, the resulting path is heavily dictated by the underlying routing algorithm. Consequently, refining the path-finding logic constitutes a high-priority objective for subsequent methodological advancements in the field.

\section*{Acknowledgement and funding}
The authors thank Cuebiq Inc. for sharing the GPS dataset used in this work.
\\
The authors acknowledge the support from MUR, grant Dipartimento di Eccellenza 2023-2027.

\section*{Disclosure statement}
The authors report that there are no competing interests to declare.

\bibliographystyle{IEEEtran}
\bibliography{bibliography}

\end{document}